\documentclass{article}

\usepackage[preprint,nonatbib]{preprint_2026}
\usepackage[utf8]{inputenc}
\usepackage[T1]{fontenc}
\usepackage{hyperref}
\hypersetup{
  colorlinks=true,
  citecolor=black,
  linkcolor=black,
  urlcolor=black,
  hypertexnames=false,
  pdftitle={RCWT: Measuring Task-Budget Displacement from Coordination Content in LLM Calls},
  pdfauthor={Brenda Lelis and Rodrigo Cabral-Carvalho}
}
\usepackage{url}
\usepackage{booktabs}
\usepackage{amsfonts}
\usepackage{amsmath,amssymb}
\usepackage{nicefrac}
\usepackage{microtype}
\usepackage{xcolor}
\usepackage{graphicx}
\usepackage{float}
\usepackage{array}
\providecommand{\tightlist}{\setlength{\itemsep}{0pt}\setlength{\parskip}{0pt}}

\title{RCWT: Measuring Task-Budget Displacement from Coordination
Content in LLM Calls}
\author{Brenda Lelis\thanks{Corresponding author.}\\
CloudWalk, Inc.\\
São Paulo, Brazil\\
\texttt{brenda.lelis@cloudwalk.io}
\And
Rodrigo Cabral-Carvalho\\
CloudWalk, Inc.\\
São Paulo, Brazil\\
\texttt{rodrigo.motta@cloudwalk.io}}

\begin{document}
\maketitle

\begin{abstract}
Multi-agent and memory-augmented LLM systems often place coordination content, shared state, prior discussion, tool outputs, summaries, and role instructions, inside the same finite prompt used for the current task. This creates a practical allocation problem: every token spent on coordination is unavailable to task instructions or evidence when a call is assembled under a fixed context budget. We introduce the Roundtable Context Window Test (RCWT), a controlled protocol for measuring this task-budget displacement effect. RCWT varies coordination content while controlling total budget, position order, task family, and scoring. In the main context-dependent recall task at $W=4096$, three commercial models remain near baseline through moderate overhead and then degrade sharply once residual reference evidence falls to a few hundred tokens. Window-scaling summaries are consistent with a task-specific residual-budget interpretation rather than a fixed percentage threshold, but we treat this as descriptive evidence rather than a universal law. To test whether the fixed-budget cliff persists when task evidence remains intact, we add an intact-task ablation: the full task/reference block is kept present while coordination tokens increase by expanding total prompt length. In that setting, all tested calls return every scored field correctly across GPT-4.1-mini, Claude Haiku 4.5, and Gemini 2.5 Flash up to a 95\% coordination ratio. This ablation narrows the claim: the main RCWT cliff is best read as task-budget displacement, not as proof that coordination volume alone causes semantic interference in the original open-ended task. RCWT is therefore a measurement primitive for context-allocation budgeting, not a complete theory of multi-agent benefit or session-level coordination.
\smallskip

\end{abstract}

\section{Introduction}\label{introduction}

Modern LLM applications assemble many kinds of information into a single
call: task instructions, retrieved documents, memory summaries, prior
agent messages, tool observations, role prompts, and shared state
\cite{ram-etal-2023-context,toolformer,autogen,guo2024multiagents}. The
application-level question is not only whether a model advertises a
large context window, but how much of the current call is allocated to
the task block that must be read and acted on. Prior long-context work
shows that models do not use all positions uniformly and that longer
inputs can degrade performance even when relevant information is
available \cite{lostmiddle,longbench,ruler,du2025contextlength}.

This paper studies a narrow, operational version of that problem. For a
single model call with context budget \(W\), let \(c\) be coordination
content and let \(W-c\) be the remaining task block. We estimate

\[
R_M(c,W,T) \in [0,1],
\]

where \(M\) is the model and \(T\) is the task family. We use the term
\textbf{task-budget displacement} for the main effect: under fixed
\(W\), increasing coordination content can physically reduce or truncate
the task evidence needed by the current call.

We introduce the Roundtable Context Window Test (RCWT). RCWT is a
single-call benchmark protocol that varies coordination allocation while
controlling prompt order and scoring. It intentionally excludes full
multi-agent session dynamics such as turn scheduling, retrieval policy,
memory writes, tool failures, and agent topology. Those factors matter,
but mixing them into the same experiment would obscure the local
allocation effect.

The contributions are:

\begin{enumerate}
\def\labelenumi{\arabic{enumi}.}
\tightlist
\item
  \textbf{A controlled protocol.} RCWT varies coordination allocation
  under fixed budget with position control, explicit token accounting,
  and task-level scoring.
\item
  \textbf{A fixed-budget displacement result.} On a
  technical-specification recall task, accuracy remains high at moderate
  overhead and drops sharply only when the residual task block becomes
  very small.
\item
  \textbf{A truncation-disambiguating ablation.} When the full task
  block remains intact and total prompt length grows to accommodate
  coordination, we detect no cliff-sized degradation across tested
  models and ratios.
\item
  \textbf{Task dependence and boundary evidence.} Self-contained tasks
  remain stable, while passage-heavy benchmark packs require larger
  residual task budgets and expose a model--task exception before
  truncation.
\item
  \textbf{A scoped engineering implication.} Coordination context should
  be budgeted against task-specific residual needs. RCWT does not
  measure the net benefit of coordination.
\end{enumerate}

\section{Related Work}\label{related-work}

\textbf{Long-context behavior.} The Lost in the Middle result showed
that LLMs can fail to use relevant information depending on position
\cite{lostmiddle}. LongBench and RULER broaden long-context evaluation
across retrieval, QA, summarization, synthetic tasks, and configurable
sequence lengths \cite{longbench,ruler}. RCWT differs by holding a call
budget fixed and varying allocation between coordination-like content
and task evidence.

\textbf{Attention and systems bottlenecks.} Long-context use is
constrained by attention and KV-cache costs. LongLoRA, FlashAttention,
Longformer, and BigBird address feasibility through training or
attention mechanisms \cite{longlora,flashattention,longformer,bigbird}.
RCWT asks a complementary application-level question: given a feasible
context, how much of it is left for the task block?

\textbf{Context compression.} Prompt compression systems such as
LLMLingua and LLMLingua-2 reduce token count while trying to preserve
task-relevant information \cite{jiang-etal-2023-llmlingua,llmlingua2}.
RCWT can evaluate whether compression moves a call away from a
high-displacement regime.

\textbf{Multi-agent coordination.} Multi-agent LLM systems introduce
communication, role, orchestration, verification, and memory overhead
\cite{mast,scalingagents}. They may also produce benefits through
decomposition, critique, and specialized roles. RCWT measures only the
per-call token cost side of that ledger. It does not estimate net system
value.

\section{Method}\label{method}

\subsection{Fixed-budget RCWT}\label{fixed-budget-rcwt}

The main RCWT task is context-dependent recall over a technical software
specification. The task asks for a structured technical analysis whose
scored facts must be recovered from a reference block. The coordination
block is synthetic but structured like multi-agent shared context:
role/protocol text, agent messages, shared propositions, and tool
schemas.

Token accounting is deterministic in the artifact. Prompt construction
uses the \texttt{cl100k\_base} tokenizer to size blocks. Let \(u\) be
the fixed task-instruction block, \(c\) the coordination block, and
\(r\) the technical-reference block, so that \(c+u+r=W\). In the
released main runs, \(u=337\) construction tokens. The runner controls a
target ratio \(q=c/(W-u)\), then sets \(c=\lfloor q(W-u)\rfloor\) and
\(r=W-u-c\). The full-budget coordination share used in the analysis is
therefore \(p=c/W\), and the residual task budget is \(W-c=u+r\). The
raw \texttt{proportion} field stores \(q\), not \(p\).

For each target allocation, the coordination template is repeated or
prefix-truncated to \(c\) tokens. The reference template is likewise
repeated or prefix-truncated to \(r\) tokens. Provider-reported token
counts are saved with every trial, but the allocation itself is built
from this common tokenizer so the sweep is reproducible across
providers.

The main experiment uses \(W=4096\) and target ratios

\[q \in \{0,25,50,75,90,92,94,96,98\}\%.\]

These targets realize
\(p\in\{0,22.9,45.9,68.8,82.6,84.4,86.3,88.1,89.9\}\%\) after accounting
for \(u\). Each condition is run in two orders: \texttt{coord\_first}
and \texttt{reason\_first}. The primary suite uses \(N=20\) trials per
order, or 40 calls per model--target pair. The tested model IDs are
\texttt{gemini-2.0-flash}, \texttt{claude-haiku-4-5-20251001}, and
\texttt{gpt-4.1-mini}. The original open-ended responses are scored with
a binary fact-checking judge over 10 items, with two known floor-effect
items excluded from the effective score. The 8 effective items are:

\begin{enumerate}
\def\labelenumi{\arabic{enumi}.}
\tightlist
\item
  three encryption options;
\item
  recommendation of per-document symmetric key;
\item
  PostgreSQL LISTEN/NOTIFY 8KB payload limit;
\item
  Redis Pub/Sub approximately 100K messages/sec;
\item
  Redis Streams approximately 5ms latency versus Pub/Sub approximately
  1ms;
\item
  ProseMirror as the rich-text recommendation;
\item
  backend team needs 1 week CRDT ramp-up;
\item
  WebSocket/infrastructure migration takes about 2--3 weeks.
\end{enumerate}

The two excluded items are \texttt{crdt\_throughput} and
\texttt{mls\_rfc}, which behaved as floor-effect parsing items in the
original baseline. The artifact reports both raw and effective scores.

\subsection{Candidate curves and residual-budget
summary}\label{candidate-curves-and-residual-budget-summary}

For the main task, we fit logistic, power-law, exponential, quadratic,
and piecewise-linear descriptive curves. The logistic form is

\[
R_M(c,W,T)=\frac{R_0}{1+\exp(k(c/W-p_0))}.
\]

This is not a mechanistic proof. The cliff-region sampling makes a
threshold-like family likely to fit well. We use the curve as compact
interpolation, not as evidence of a universal law.

We also summarize the transition with a task-specific residual budget
parameter \(\theta\):

\[
p_0(W)=1-\frac{\theta}{W}.
\]

Here \(p_0\) is expressed as the realized full-budget share \(c/W\), so
\(\theta=W(1-p_0)\) is the fitted residual task budget \(u+r\); the
corresponding reference-only reserve is \(\theta-u\). This relation is
an empirical description over the tested window sizes. With only a small
number of window sizes, \(\theta\) should be read as a calibration
estimate, not as a validated invariant.

\subsection{Intact-task ablation}\label{intact-task-ablation}

We add an ablation that probes whether a cliff-sized degradation
persists when task evidence is kept intact rather than truncated. The
full task/reference block is kept intact in every condition.
Coordination tokens are varied around it, and the total prompt length
grows accordingly. Thus, the task is never physically shortened.

The ablation uses deterministic JSON scoring rather than an LLM judge.
It asks for eight exact fields matching the same reference facts used in
the main task. In this ablation, the coordination ratio is \(c/(c+t)\),
where \(t\) is the intact task/reference block; this differs from the
fixed-budget experiment, where the ratio is \(c/W\). The measured
task/reference block has \(t=698\) construction tokens. Tested ratios
are \(0, 0.50, 0.75, 0.90, 0.95\), corresponding to 0, 698, 2,094,
6,282, and 13,262 coordination tokens and estimated prompt sizes of 702,
1,401, 2,797, 6,985, and 13,965 tokens. Both prompt orders are tested
with \(N=5\) per order, so each model-ratio cell pools \(10\) calls and
\(80\) binary field decisions. Models are GPT-4.1-mini, Claude Haiku
4.5, and Gemini 2.5 Flash. The Gemini model differs from the historical
main table because Gemini 2.0 Flash was unavailable during later reruns.

\subsection{Boundary tasks and packs}\label{boundary-tasks-and-packs}

We additionally retain two boundary probes:

\begin{itemize}
\tightlist
\item
  \textbf{Self-contained algorithmic task:} the task block contains all
  information needed to answer execution-trace questions; coordination
  is irrelevant.
\item
  \textbf{External benchmark packs:} GSM8K, MMLU-Pro, and DROP
  \cite{gsm8k,mmlupro,drop} are grouped into 10-item packs so that
  task-block size can become binding under fixed \(W\). Each
  model--share aggregate contains 100 calls: 10 items, two prompt
  orders, and five repeated trials.
\end{itemize}

\section{Results}\label{results}

\subsection{Fixed-budget main task}\label{fixed-budget-main-task}

The main fixed-budget result shows stable accuracy through moderate
overhead, followed by sharp degradation when the task block becomes very
small.

\begin{figure}[H]
\centering
\includegraphics[width=\linewidth]{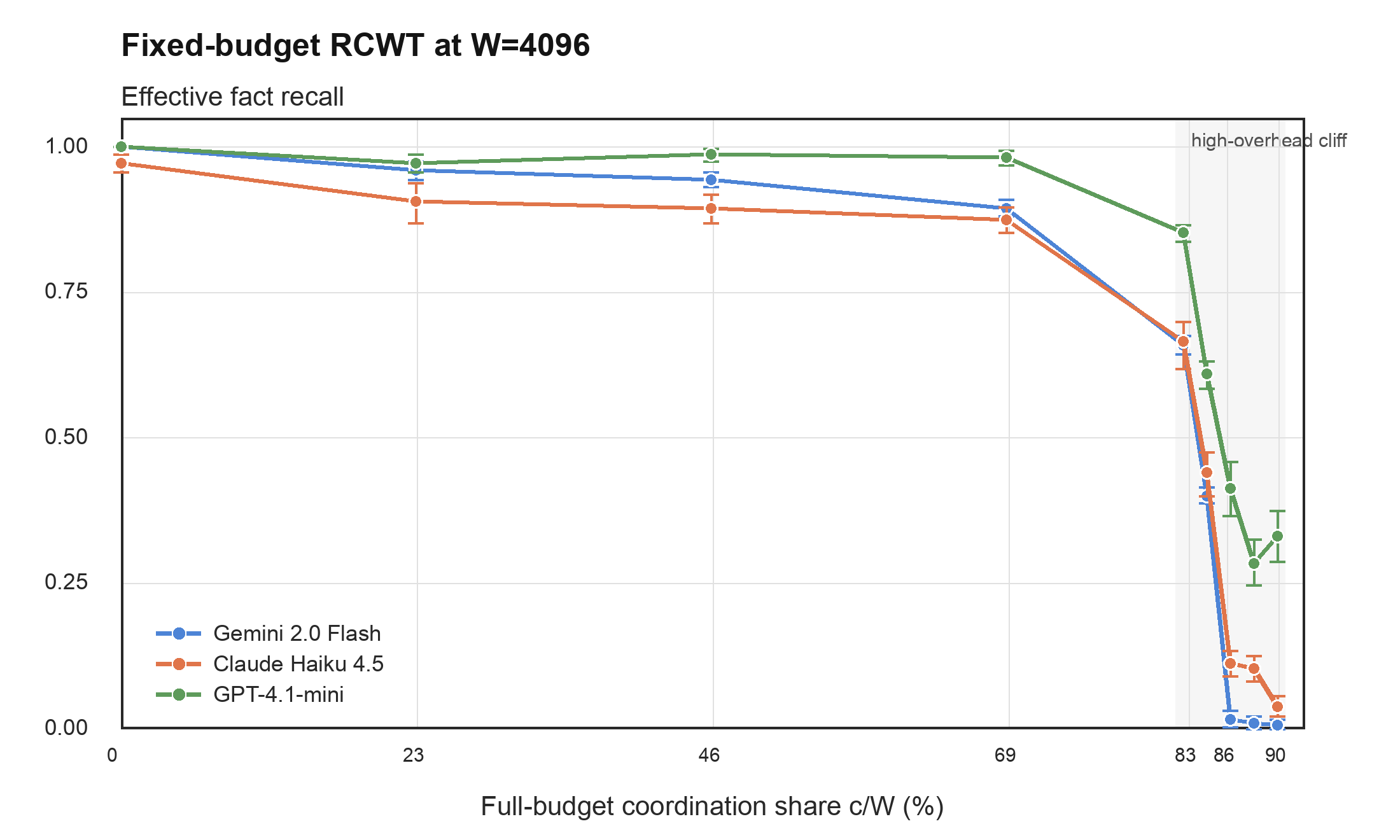}
\caption{Fixed-budget RCWT at $W=4096$ across three providers. The horizontal axis reports the realized full-budget coordination share $p=c/W$. Error bars are 95\% stratified call-level bootstrap intervals over the two prompt orders. Effective fact recall remains high through moderate coordination overhead, then falls sharply once residual reference evidence is reduced to a few hundred tokens.}
\label{fig:rcwt-cross-provider}
\end{figure}

Figure \ref{fig:rcwt-cross-provider} shows the main fixed-budget result.
All three models remain close to baseline at moderate coordination
overhead, then drop once residual reference evidence is reduced to a few
hundred tokens. Exact aggregate values and the distinction between
target ratio \(q\) and realized share \(p=c/W\) are reported in Appendix
Table \ref{tab:main-fixed-budget-values}.

At target \(q=90\%\), a realized coordination share of
\(p=82.6\%\), the reference block contains 376 tokens and the total
residual task budget \(u+r\) is 713 tokens. The drop from baseline is
34.1 percentage points for Gemini, 30.6 for Haiku, and 14.7 for GPT.
This supports a fixed-budget warning: average prompt length and nominal
context window are not sufficient safety signals; the residual task
budget matters.

\subsection{Window scaling}\label{window-scaling}

A fixed-percentage account would place the transition at roughly the
same realized \(p\) across window sizes. The observed summaries do not
match that account. At target \(q=90\%\), \(W=4096\) realizes
\(p=82.6\%\) and leaves 376 reference tokens; \(W=8192\) realizes the
larger share \(p=86.3\%\) yet leaves 786 reference tokens and shows less
degradation. The \(W=8192\) sweep uses 20 trials per order (40 calls per
model--target pair); the \(W=16384\) sweep uses 10 per order (20 calls
per pair).

\begin{table}[H]
\centering
\small
\caption{Window-scaling summary for the main task. $\theta$ is estimated from the $W=4096$ fit. This is descriptive calibration, not a validated invariant.}
\begin{tabular}{lrrrr}
\toprule
Model & $\theta$ & Pred. $p_0$(8K) & Obs. $p_0$(8K) & Obs. $p_0$(16K) \\
\midrule
Gemini 2.0 Flash & 665 & 91.9\% & 92.1\% & 95.9\% \\
Claude Haiku 4.5 & 650 & 92.1\% & 92.2\% & 96.0\% \\
GPT-4.1-mini & 568 & 93.1\% & 93.1\% & 96.4\% \\
\bottomrule
\end{tabular}
\end{table}

The relation is consistent with residual-budget displacement over these
runs. It is not strong evidence that \(\theta\) is stable outside this
task, these models, or these window sizes.

\subsection{Intact-task ablation}\label{intact-task-ablation-1}

The task/reference block is never truncated in this ablation.
Coordination tokens are added around the intact task block, increasing
total prompt length. Every model--ratio cell has 10/10 calls with all
eight fields correct, equivalent to 80/80 correct field decisions.
Treating complete-call success as the unit, 10/10 gives a Wilson 95\%
interval of approximately \([0.722,1.000]\) \cite{wilson1927probable}.
Across the 15 heterogeneous cells, the descriptive totals are 150/150
complete calls and 1200/1200 correct fields; we do not use the pooled
field count as an independent-sample confidence interval.

The main fixed-budget cliff does not, by itself, show that coordination
text harms reasoning while task evidence remains intact. In this task
and scoring setup, we detect no cliff-sized semantic-interference effect
even with large coordination blocks, as long as the reference remains
present. The sample does not exclude smaller effects. Because the
ablation also changes the task from open-ended structured analysis to
exact-field extraction, it is easier than the main task; it should
therefore be interpreted as evidence against a large interference effect
in this extraction-style intact-evidence setting, not as a full
semantic-interference test for the original task format. The main effect
is best interpreted as displacement or truncation of task evidence under
fixed budget.

\subsection{Boundary tasks and external
packs}\label{boundary-tasks-and-external-packs}

The self-contained algorithmic task remains stable across overhead
levels for all three providers. This shows that extra text alone is not
sufficient to cause degradation in the tested range.

Most model--pack pairs remain close to baseline while the full task pack
fits, then fall when task content is partially or fully truncated. One
material exception is Claude Haiku 4.5 on the untruncated GSM8K pack:
accuracy falls from 0.60 at \(p=0\) to 0.40 at \(p=0.25\) and 0.38 at
\(p=0.50\). The pack probes therefore support task-budget sensitivity
but do not identify truncation as the only possible source of
degradation.

\begin{table}[H]
\centering
\small
\caption{Task-dependent logistic midpoint summaries at $W=4096$. Pack probes are RCWT stress probes, not leaderboard estimates.}
\begin{tabular}{lrrl}
\toprule
Probe & Logistic $p_0$ range & Approx. reserve & Truncation onset \\
\midrule
Task 1 spec recall & 0.838--0.861 & 568--665 tok & target $q=90\%$ \\
GSM8K-pack & 0.853--0.893 & 438--602 tok & 90\% \\
MMLU-Pro-pack & 0.733--0.755 & 1003--1094 tok & 50\% partial \\
DROP-pack & 0.481--0.624 & 1539--2126 tok & 50\% \\
\bottomrule
\end{tabular}
\end{table}

\section{Discussion}\label{discussion}

\subsection{Scope of the contribution}\label{scope-of-the-contribution}

The controlled allocation protocol is useful. RCWT isolates one factor
that appears in real multi-agent systems: coordination content can
consume the same prompt budget needed for task evidence. The main data
support a practical engineering rule: measure the residual task budget
needed by a task family and budget coordination around it.

\subsection{Limits of interpretation}\label{limits-of-interpretation}

A broader interpretation, that the paper proves general reasoning
degradation from semantically irrelevant coordination, is not
supported. The main task is recall-heavy; the cliff aligns with small
residual reference blocks; the intact-task ablation stays at ceiling.
The supported claim is narrower: RCWT measures fixed-budget task
displacement, and semantic interference requires separate evidence. The
ablation provides evidence against a cliff-sized semantic effect in this
extraction-style setup; it should not be read as proof that semantic
interference is zero in the original open-ended task format.

\subsection{Cost versus benefit}\label{cost-versus-benefit}

RCWT measures cost, not net value. Real coordination can improve quality
through decomposition, verification, tool use, or memory. A coordination
block can be overhead for one agent and the task input for another. The
correct practical question is not whether coordination hurts, but
whether this coordination content provides enough benefit to justify the
task-budget it consumes. RCWT supplies only the cost-side measurement.

\subsection{Coordination
heterogeneity}\label{coordination-heterogeneity}

The central coordination block is synthetic and structured. Real
coordination varies: dense tool outputs, verbose transcripts, retrieved
documents, distilled state, uncertainty annotations, or contradictory
agent claims may behave differently at the same token count. The current
results should not be generalized across all coordination types without
a design that varies content type independently of length.

\section{Limitations}\label{limitations}

The fixed-budget sweep was developed iteratively: cliff-region cells
were added after early flat-region observations. The resulting curves
are compact descriptive summaries, not pre-registered tests of a
universal functional form. The main figure uses a stratified call-level
bootstrap over the two prompt orders. Because calls reuse the same task
and reference facts, these intervals quantify run-to-run variation
conditional on this prompt; they do not represent uncertainty over a
population of independently sampled tasks.

The main result is strongest for one technical-specification recall
task. The benchmark packs broaden coverage but do not constitute a
pre-registered task-complexity ladder. In the main task, prefix
truncation removes facts in a fixed order, so the fitted midpoint partly
reflects reference layout and fact order rather than an intrinsic task
constant. The coordination template also overlaps topically with some
scored concepts, including CRDTs, MLS, Redis, WebSockets, and migration
timing. Future runs should randomize fact order and vary coordination
content independently of topical overlap.

The main task uses an LLM judge for open-ended parsing. Cross-vendor
rescoring did not remove the qualitative cliff, but judge-specific
strictness remains a limitation. The intact-task ablation uses strict
per-field deterministic JSON scoring to reduce this risk, but it also
changes response format, task demand, and scoring method. Therefore, the
ablation is best interpreted as a test for a large semantic-interference
effect in an extraction-style intact-evidence setting, not as a
token-identical rescore of the open-ended main task. A deterministic
rescore or human audit of the main open-ended responses remains
complementary validation.

Provider tokenizers differ. The scripts use \texttt{cl100k\_base} for
construction plus provider-reported token counts where available. The
distinction between the runner target \(q=c/(W-u)\) and the reported
full-budget share \(p=c/W\) is now explicit, but native token accounting
should be preferred in future runs.

Gemini 2.0 Flash was available during the original fixed-budget runs but
was unavailable during later reruns. Historical aggregate files are
retained; new reruns and the intact-task ablation use current model IDs.
RCWT is a single-call protocol and does not model session-level
multi-agent dynamics, retrieval policy, tool failure, memory
summarization, or coordination benefit.

\section{Conclusion}\label{conclusion}

RCWT is a controlled protocol for measuring task-budget displacement
from coordination content in LLM calls. The main fixed-budget experiment
shows a sharp high-overhead cliff on a context-dependent recall task,
and window summaries are consistent with a task-specific residual-budget
interpretation. The intact-task ablation shows that when the full task
block remains present, we detect no large degradation from extra
coordination content in an extraction-style setup, while small semantic
effects and harder-task interference remain possible. This narrows the
contribution: coordination context is not cost-free under fixed budgets,
but the main evidence supports displacement of task evidence rather than
a general semantic competition law. Practical systems should track
residual task budget, not only nominal context size or total prompt
length.

\section*{Code and artifact availability}

The source code, prompt templates, and aggregate result files are
available in the public repository:
\url{https://github.com/cloudwalk/rcwt-agent-coordination}.

\section*{Acknowledgements}

LLM tools were used for grammar correction and drafting assistance. The
authors take responsibility for the experiments, analysis, and claims.

\appendix
\setcounter{table}{0}
\renewcommand{\thetable}{A\arabic{table}}
\section*{Appendix A. Prompt and task examples}

The experiments use controlled prompt fragments rather than production
logs. Full prompt templates and aggregate files are kept in the
open-source artifact repository.

\textbf{Main context-dependent task.} The task/reference block is a
technical software specification. The coordination block is synthetic
role, protocol, shared-state, and agent-message text varied
independently of that reference. Representative reference facts state
that PostgreSQL LISTEN/NOTIFY has an 8KB payload limit, Redis Pub/Sub
handles roughly 100K messages per second but has no persistence, and
Redis Streams provides persistence with about 5ms latency versus about
1ms for Pub/Sub. The model is asked for a structured technical analysis,
and binary scored items check whether these facts are recovered
correctly.

\textbf{Self-contained algorithmic task.} The task block gives a Python
function that partitions
\texttt{items={[}3,\ 10,\ 15,\ 7,\ 10,\ 22,\ 4{]}} around
\texttt{threshold=10}. Representative scored questions ask whether the
sum is 71, whether indices 2 and 5 are above threshold, and whether the
maximum above-threshold value is 22. The answers are fully determined by
the code and input, so the coordination block is irrelevant.

\textbf{External benchmark packs.} GSM8K, MMLU-Pro, and DROP examples
are grouped into 10-item packs. Each model--share aggregate contains 100
calls across 10 items, two prompt orders, and five repeated trials. The
pack probes are stress tests for task-budget sensitivity rather than
leaderboard estimates.

\begin{table}[H]
\centering
\small
\caption{Detailed main context-dependent RCWT task at $W=4096$, pooled over position order. The runner controls $q=c/(W-u)$; $p=c/W$ is the realized full-budget coordination share, $u=337$, and $r$ is the residual reference block. Scores are effective binary accuracy over 40 calls per model--target pair.}
\label{tab:main-fixed-budget-values}
\resizebox{\linewidth}{!}{%
\begin{tabular}{lrrrrr}
\toprule
Target $q$ & Realized $p$ & Ref. $r$ & Gemini 2.0 Flash & Haiku 4.5 & GPT-4.1-mini \\
\midrule
0\%  & 0.0\%  & 3759 & 1.000 & 0.972 & 1.000 \\
25\% & 22.9\% & 2820 & 0.960 & 0.906 & 0.972 \\
50\% & 45.9\% & 1880 & 0.944 & 0.894 & 0.988 \\
75\% & 68.8\% & 940  & 0.894 & 0.875 & 0.981 \\
90\% & 82.6\% & 376  & 0.659 & 0.666 & 0.853 \\
92\% & 84.4\% & 301  & 0.400 & 0.441 & 0.609 \\
94\% & 86.3\% & 226  & 0.016 & 0.113 & 0.413 \\
96\% & 88.1\% & 151  & 0.009 & 0.103 & 0.284 \\
98\% & 89.9\% & 76   & 0.006 & 0.038 & 0.331 \\
\bottomrule
\end{tabular}
}
\end{table}

\end{document}